%% file: root.tex
\documentclass{article}

\usepackage[preprint]{corl_2026} 
\usepackage{graphicx}
\usepackage{booktabs,multirow,makecell,xcolor, placeins,caption, amsfonts}
\title{Editing the Goal: Per-Token SAE Steering and Its Depth Limits in $\pi_{0.5}$}
\author{
  Rocio Perales Valdes\\
  PAIR Lab, AI Safety Initiative\\
  Georgia Institute of Technology 
  United States\\
  \texttt{rocio.perales@gatech.edu} \\
   \And
  Animesh Garg\\
  PAIR Lab\\
  Georgia Institute of Technology 
  United States\\
  \texttt{animesh.garg@gatech.edu} \\
}

\begin{document}
\maketitle


\input{sections/abstract}


\input{sections/introduction}
\label{sec:introduction}

\input{sections/Method}
\label{sec:method}
	

\input{sections/experimental_results}
\label{sec:results}


\input{sections/conclusion}
\label{sec:conclusion}


\clearpage


\bibliography{example}  

\end{document}

%% file: sections/abstract.tex
\begin{abstract}
Sparse autoencoders (SAEs) have recovered interpretable, causally relevant features in robot foundation models (RFMs) such as $\pi_{0.5}$~\cite{swann2026drvla,grant2026features}. SAE features on individual instruction tokens are less interpretable than timestep-pooled ones~\cite{swann2026drvla}, and whether they can redirect a robot's goal remains untested. We show that replacing the SAE features of a single goal word with those of an alternative goal word, leaving the instruction text unchanged, steers $\pi_{0.5}$-LIBERO into completing the alternative task. At layer 0, BatchTopK feature swaps redirect behavior in up to 20/20 rollouts in a scene where random-feature swaps redirect none, with success varying across six scenes in two LIBERO suites and across layers. The effect depends on SAE architecture: word-selected Matryoshka edits largely do not outperform random ones. SAEs are also not required. Adding a fixed difference-in-means vector at twice its natural magnitude redirects 8--10/10 rollouts. Layer-wise activation patching shows that goal-word edits remain effective up to layer 6 for object location and layer 2 for object identity, and up to layer 13 in both cases when patching also covers camera tokens. These results are proof of existence in an RFM in simulation and an initial map of how deep goal-redirecting edits remain effective. Visit \href{https://doubleblind-sae-pi05.github.io/}{\texttt{doubleblind-sae-pi05.github.io}} for video rollouts.
\end{abstract}

\keywords{Sparse Autoencoders, Vision-Language-Action Models, Activation Steering, Mechanistic Interpretability}

\begin{center}
    \includegraphics[width=0.8\linewidth]{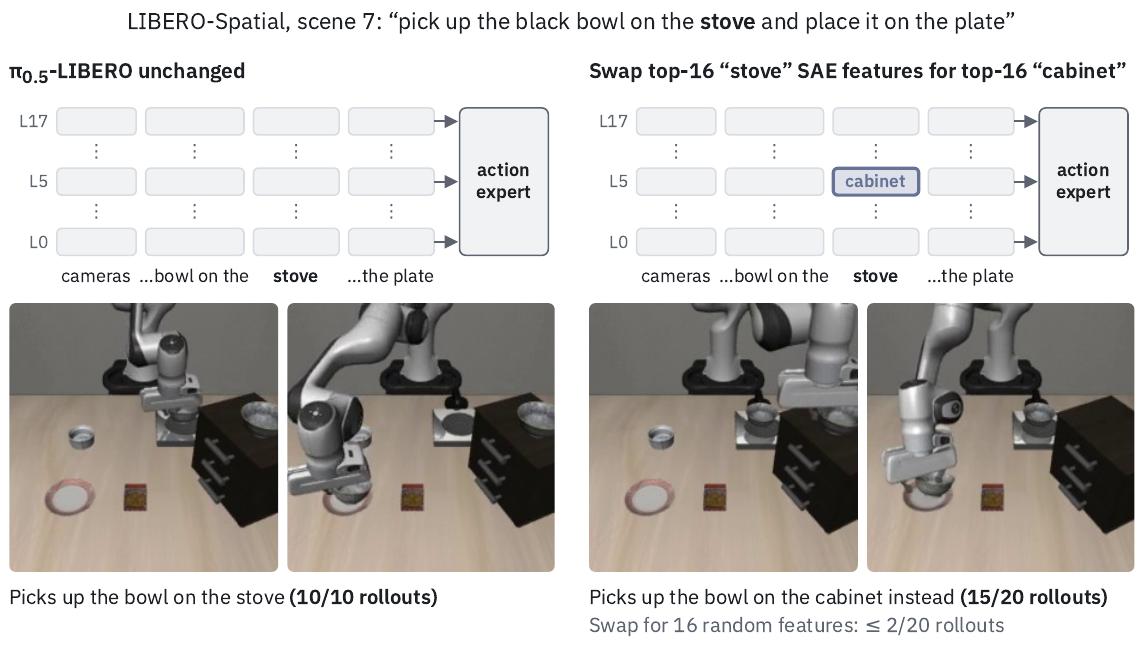}
    \captionof{figure}{Swapping one word's SAE features at layer 5 changes the robot's goal in 15/20 rollouts. Both runs see scene 7 and read the same instruction, but they pick different bowls.}
    \label{fig:swap}
\end{center}

%% file: sections/introduction.tex
\section{Introduction}
Studying the steerability and performance of vision-language-action models (VLAs) is essential as dangerous trajectories in the physical world may be irreversible or have expensive effects. As such, transparency in the goal a model is pursuing and ensuring it has not been tampered with is needed for extensive use. Prior work steers VLAs along sparse or feed-forward directions~\cite{khan2025sparse,haon2025steering} or with linear observers and controllers~\cite{buurmeijer2026observing}. Li~\cite{li2025textlatent} shows that hidden states of instruction tokens, averaged over a task's demonstrations (a ``text latent''), can be written into the residual stream of $\pi_0$~\cite{black2025pi0} to reconstruct tasks, and combined to enable new capabilities without retraining.

\citet{swann2026drvla} find interpretable, causal SAE features in $\pi_{0.5}$, mostly from activations pooled over each timestep, and \citet{grant2026features} find that steering per-token SAE features in $\pi_{0.5}$'s action expert mostly degrades the policy rather than redirecting it. Whether a single instruction word's SAE features can redirect the goal, and how deep an instruction can be edited, remain open.

In this work, we (i) show that swapping a goal word's per-token SAE features redirects $\pi_{0.5}$-LIBERO to another instruction's task in up to 20/20 rollouts while preserving success, with feature combinations succeeding where single features fail; (ii) compare feature-selection methods (word-selected and contrastive), SAE architectures (BatchTopK and Matryoshka) and non-SAE baselines (activation patching and difference vectors); and (iii) map how deep instructions stay editable by patching one layer at a time for the goal word, the instruction, and the instruction with the camera tokens.

%% file: sections/method.tex
\section{Method}
\textbf{SAEs.} We train two SAE architectures on the PaliGemma~\cite{beyer2024paligemma} residual stream of $\pi_{0.5}$-LIBERO~\cite{intelligence2025pi05,liu2023libero}, using the openpi checkpoint~\cite{openpi2025}, at layers 0, 5, 11 and 17. ``Layer $L$'' is the output of PaliGemma block $L$. Each SAE input (row) is either the mean of one camera's 256 image tokens or an individual text token, following the per-token setup of~\citet{swann2026drvla}. We train BatchTopK SAEs~\cite{bussmann2024batchtopk} with the AuxK loss of~\citet{gao2024scaling} and the settings of~\citet{swann2026drvla}, who use per-row TopK at $1\times$ expansion (2,048 latents). We also train Matryoshka SAEs with groups of 64, 64, 128, 256, 512 and 1024 latents, under the intuition that their reduced feature splitting and hierarchical structure~\cite{bussmann2025matryoshka} would transfer to RFMs. All trained SAEs reconstruct $>98\%$ of the variance.
\newline\textbf{Scene selection.} LIBERO policies often ignore the instruction~\cite{zhou2025liberopro, grant2026features}: in LIBERO-Spatial, $\pi_{0.5}$-LIBERO with no prompt succeeds in 61/100 episodes. We only use scenes where a null prompt leads to $\leq 3/10$ successes. In LIBERO-Spatial (10 tasks), the policy must pick the correct one of two black bowls and place it on the plate. Scenes 7 (``on the stove'') and 9 (``on the wooden cabinet'') form a mirror pair. We use this pair and cross-check the results on LIBERO-Object, where the instructions differ only in the object to be placed in the basket. There we pick four scenes where changing the instruction redirects the policy in 5/5 rollouts. The two suites also let us compare how deep goals stay editable for two classes of instructions (object location vs.\ object identity).

\textbf{Steering.} Let $h^{\ell}_i(p)$ be the residual stream at the output of block $\ell$ for token $i$ under prompt $p \in \{p_A, p_B\}$, and $T_A$, $T_B$ be the goal-word tokens of each prompt (one or two per word, e.g.\ ``wooden cabinet''); $\bar{h}_T$ averages $h$ over $T$. All edits target the goal word of $p_A$, leave the instruction text unchanged, and are reapplied at every policy call.

\textit{SAE swap.} For each $i \in T_A$, we encode $\mathbf{x} = h^L_i(p_A)$ into codes $\mathbf{z}$, set $z_j$ to a target value $c_j$ for every latent $j$ in a set $\mathcal{S}$ to get $\mathbf{z}'$, and add the change back through the decoder, $\mathbf{x}' = \mathbf{x} + \mathrm{Dec}(\mathbf{z}') - \mathrm{Dec}(\mathbf{z})$, keeping the reconstruction error. Due to the SAE rescaling each row to unit norm, we re-encode eight times, bringing the edited codes to within 0.18 of their targets. In the re-encoded rows of word-selected BatchTopK swaps, a median of 46 non-edited latents switch on or off at layer 5 and 397 at layer 0, against 61 and 124 when A's word is replaced by B's. We choose $\mathcal{S}$ in two ways. \emph{Word-selected:} for each of $T_A$ and $T_B$, the 16 latents (or all eligible ones, up to 64) most selective for that word, i.e.\ with the highest ratio of mean activation on the word, over every instruction containing it, to mean activation on all other text tokens. We skip latents active on more than half of all rows, such as always-on attention-sink latents~\cite{xiao2023sinks}; $T_A$'s latents are set to $c_j = 0$ and $T_B$'s to their mean activation on $T_B$. \emph{Contrastive:} the 16 (or 64) latents whose codes at the goal word change most between prompts A and B on the same demonstration frames, with $c_j$ their mean under prompt B.

\textit{Difference vector.} Following~\citet{rimsky2024caa}, we add $\alpha v$ to the goal-word rows, where $v$ is the mean over 256 demonstration frames of task A of $\bar{h}^L_{T_B}(p_B) - \bar{h}^L_{T_A}(p_A)$, so $\alpha = 1$ adds the mean difference itself. Unlike patching, it needs no prompt-B pass at test time.

\textit{Activation patching.} We set $h^L_i(p_A) \leftarrow \bar{h}^L_{T_B}(p_B)$ for $i \in T_A$, with prompt B's states taken from a separate pass on the same frame, and recompute all later blocks. The goal word is patched either once at $L$ or at every block from $L$ onward. To analyze whether the rest of the instruction and the camera tokens also carry the goal, we patch incrementally, first adding the whole instruction and then the instruction and camera tokens, each set to its own state under prompt B from $L$ onward. Tokens before $T$ are matched by position and tokens after it from the end of the instruction, so each word has a counterpart even when lengths differ (``stove'' vs.\ ``wooden cabinet'').

\textit{Single-feature ablation.} We also remove one latent's decoder direction from every prefix token, camera and text, adapting the all-token intervention of~\citet{swann2026drvla}.

\textbf{Success and controls.} All conditions in a scene share initial state and seed, and success is awarded if B's task is completed. Each steering experiment also has a control edit: for SAE swaps, the same number of random latents set to the same target values $c_j$ (three draws; with word selection they replace only $T_B$'s latents, and $T_A$'s are still set to zero); for the difference vector, random vectors of the same norm (three seeds); for activation patching, an arbitrary non-target word of $p_B$.
\input{tables/table1}

%% file: tables/table1.tex
\begin{table}[b]
\centering\footnotesize
\setlength{\tabcolsep}{3pt}
\begin{tabular}[t]{@{}lcccc@{}}
\multicolumn{5}{@{}l}{\textbf{(a)} LIBERO-Spatial (of 20)}\\
\toprule
 & \multicolumn{2}{c}{BatchTopK} & \multicolumn{2}{c}{Matryoshka}\\
\cmidrule(lr){2-3}\cmidrule(l){4-5}
Scene & 16, $k$ & \textit{rand.} & 16, $k$ & \textit{rand.}\\
\midrule
\multicolumn{5}{@{}l}{\textit{Layer 0, word-selected}}\\
7: stove $\to$ cabinet & \textbf{20}, 8 & \textit{0} & 8, 7 & \textit{0--5}\\
9: cabinet $\to$ stove & 8, \textbf{20} & \textit{0} & \textbf{19}, 8 & \textit{1--15}\\
\multicolumn{5}{@{}l}{\textit{Layer 0, contrastive}}\\
7: stove $\to$ cabinet & 0, 5 & \textit{0} & 14, 11 & \textit{3--6}\\
9: cabinet $\to$ stove & 0, \textbf{19} & \textit{0} & \textbf{20}, \textbf{20} & \textit{4--7}\\
\midrule
\multicolumn{5}{@{}l}{\textit{Layer 5, word-selected}}\\
7: stove $\to$ cabinet & \textbf{15}, 11 & \textit{0--2} & 7, 7 & \textit{0--1}\\
9: cabinet $\to$ stove & 4, 2 & \textit{0--1} & 2, 2 & \textit{1--3}\\
\multicolumn{5}{@{}l}{\textit{Layer 5, contrastive}}\\
7: stove $\to$ cabinet & 3, 7 & \textit{0} & 0, 5 & \textit{0}\\
9: cabinet $\to$ stove & 6, \textbf{18} & \textit{0} & 5, \textbf{19} & \textit{0--2}\\
\bottomrule
\end{tabular}\hfill
\begin{tabular}[t]{@{}lcccc@{}}
\multicolumn{5}{@{}l}{\textbf{(b)} LIBERO-Object (of 20)}\\
\toprule
 & \multicolumn{2}{c}{BatchTopK} & \multicolumn{2}{c}{Matryoshka}\\
\cmidrule(lr){2-3}\cmidrule(l){4-5}
Scene & 16, $k$ & \textit{rand.} & 16, $k$ & \textit{rand.}\\
\midrule
\multicolumn{5}{@{}l}{\textit{Layer 0, word-selected}}\\
0: soup $\to$ dressing & \textbf{20}, 0 & \textit{0--1} & 13, \textbf{19} & \textit{0--15}\\
1: cream cheese $\to$ soup & \textbf{19}, 0 & \textit{5--14} & 13, 8 & \textit{11--13}\\
4: ketchup $\to$ BBQ sauce & \textbf{20}, 10 & \textit{0--1} & \textbf{15}, 7 & \textit{4--7}\\
6: butter $\to$ tomato sauce & \textbf{16}, \textbf{19} & \textit{3--4} & 10, 9 & \textit{10--17}\\
\multicolumn{5}{@{}l}{\textit{Layer 0, contrastive}}\\
0: soup $\to$ dressing & 0, 10 & \textit{0} & 0, \textbf{16} & \textit{0--7}\\
1: cream cheese $\to$ soup & 0, 0 & \textit{0} & 9, 12 & \textit{3--9}\\
4: ketchup $\to$ BBQ sauce & 0, \textbf{16} & \textit{0} & 0, \textbf{16} & \textit{0--1}\\
6: butter $\to$ tomato sauce & 2, 9 & \textit{0} & 14, 13 & \textit{6--8}\\
\midrule
\multicolumn{5}{@{}l}{\textit{Layer 5, word-selected}}\\
highest over scenes & 3, 4 & \textit{0} & 0, 1 & \textit{0}\\
\multicolumn{5}{@{}l}{\textit{Layer 5, contrastive}}\\
highest over scenes & 0, 1 & \textit{0} & 0, 0 & \textit{0}\\
\bottomrule
\end{tabular}
\caption{Rollouts (of 20) that complete prompt B's task after an SAE edit at A's goal word at layer $L$ (scene: A's goal $\to$ B's). 16, $k$: latents per word; $k$: all eligible (up to 64) for word selection, 64 for contrastive. \textit{rand.}: range over three draws of 16 random latents per word. Unedited policy with prompt B completes task B: 20/20 (a), 19--20/20 (b). Bold: $\geq 75\%$.}
\label{tab:swap}
\end{table}

%% file: sections/experimental_results.tex
\section{Results}
\textbf{Per-token SAEs enable performance-preserving steering.} At layer 0, one BatchTopK feature set makes the robot complete prompt B's task in 20/20 rollouts while instructed with prompt A. We call an edit performance-preserving when the edited policy completes B's task about as often as the unedited policy does when instructed with B. Which feature set $\mathcal{S}$ works best varies by scene and layer (Table~\ref{tab:swap}). Zeroing A's word latents without writing B's redirects at most 1/20 rollouts in the five scenes where the 16-latent layer-0 BatchTopK swap reaches $\geq$16/20, and 89\% of rollouts in Table~\ref{tab:swap} that miss B's task complete A's. High random rates (up to 17/20) come mostly from Matryoshka, whose random layer-0 edits are a median 16--54 times the size of the real word change.
 Matryoshka SAEs split words less than BatchTopK. At layer 5, we count pairs of latents where each fires on at least half of the instruction tokens the other fires on. Matryoshka has 1,323 such pairs and BatchTopK has 3,853 (0.65 vs.\ 1.88 per live latent). Contrary to our intuition, this did not help steering. On LIBERO-Object, Matryoshka's word-selected swaps do about as well as random ones.

\textbf{Single features.} Single goal-word features show little causal effect when ablated: of 40 single-feature ablations at layer 0 (the top 8 goal-word features per SAE and scene, plus 8 random), 39 keep the original task's success at $\geq 80\%$. The exception, a BatchTopK ``stove''-related feature as active on the camera rows as on the word, drops scene 7 to $<20\%$ but leaves scene 9 at 90\%; ablating it only at the camera tokens gives 6/10 and only at the word 9/10.

\textbf{SAEs are one of many viable tools.} Replacing the goal word's state with its state under prompt B leads to task B completion in 10/10 rollouts at layers 0 and 5 in both LIBERO-Spatial scenes. This follows the same-scene injection of~\citet{grant2026features} but is narrower, substituting only the goal word instead of whole layers. Alternatively, a fixed difference vector, which needs no forward pass with prompt B at test time, succeeds 5--9/10 at $1\times$ and 8--10/10 at $2\times$. Random vectors, at both $1\times$ and $2\times$ fail to redirect (0/10 success). In language models, too, simple baselines have been shown to match or outperform SAE-based steering~\cite{wu2025axbench}. In this case, the SAE adds a per-word swap (Li's text latents~\cite{li2025textlatent} are per task) that never runs prompt B on A's frames, which patching does at every policy call and the difference vector has to have done once.

\textbf{Depth of editability.} Our SAEs cover only four layers, so we use raw activation patches to study more cheaply and thoroughly how deep within the PaliGemma layers instructions remain editable (Figure~\ref{fig:depth}). The goal word stays editable (dashed blue) through layer 6 for locations but only through layer 2 for objects; holding the patch from $L$ to 17 (solid blue) and patching the whole instruction (orange) drop off very similarly. Patching the camera tokens as well still transfers the task in both suites up to layer 13 (green). An edit at layer $L$ reaches the action expert only through the keys and values of blocks $L{+}1$ to 17, so layer-17 edits cannot act, and a patch held from $L$ to 17 equals a KV-cache patch from $L{+}1$. SAE swaps work at layer 0, one block past the input embedding, partly at layer 5 (at most 4/20 in LIBERO-Object), and not at layer 11.

\begin{center}
    \includegraphics[width=0.7\linewidth]{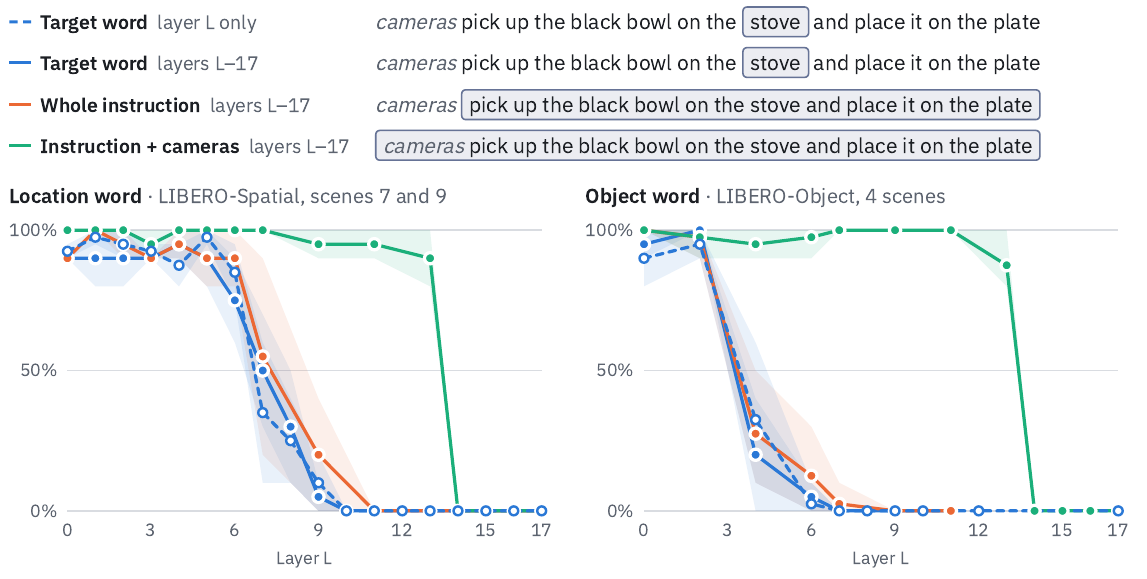}
    \captionof{figure}{Share of rollouts completing B's task after patching at layer $L$ (10 per scene, pooled; 20 for dashed blue at $L \leq 6$, left, and $L \in \{0, 2\}$ in two scenes, right). Bands: scene range.}
    \label{fig:depth}
\end{center}

%% file: sections/conclusion.tex
\section{Conclusion}
These results show that per-token SAEs can yield causal features, despite being more challenging to find: swapping one word's features at layer 0 or 5 moves the robot to another instruction's task as well as a plain difference vector. They also map the editability of instructions in RFMs, providing a cutoff, for edits at natural magnitudes, for the layers at which latent goal hijacking is a concern. Edits stay effective up to layer 6 (location) or 2 (object) at the instruction tokens and up to layer 13 across the instruction and camera tokens for one fine-tuned VLA ($\pi_{0.5}$) in simulation.